\def\paperID{0000}
\def\confName{CVPR}
\def\confYear{2025}

\def\paperTitle{MFP-VTON: Enhancing Mask-Free Person-to-Person Virtual Try-On via Diffusion Transformer}

\def\authorBlock{
    Le Shen$^{1}$ \qquad
    Yanting Kang \qquad
    Rong Huang$^{1}$ \qquad
    Zhijie Wang$^{1}$ \\
    $^{1}$Donghua University, Shanghai, China \\
    {\tt\small \{le.shen\}@mail.dhu.edu.cn \{rong.huang, wangzj\}@dhu.edu.cn}
}

\newif\ifreview 
\newif\ifarxiv \newcommand{\arxiv}{\arxivtrue}
\newif\ifcamera 
\newif\ifrebuttal 

\arxiv
\pdfoutput=1
\documentclass[10pt,twocolumn,letterpaper]{article}
\ifreview \usepackage[review]{cvpr} \fi
\ifarxiv \usepackage[pagenumbers]{cvpr} \fi
\ifrebuttal \usepackage[rebuttal]{cvpr} \fi
\ifcamera \usepackage{cvpr} \fi

\usepackage{graphicx}	
\usepackage{amsmath}	
\usepackage{amssymb}	
\usepackage{booktabs}
\usepackage{times}
\usepackage{microtype}
\usepackage{epsfig}
\usepackage{caption}
\usepackage{float}
\usepackage{placeins}
\usepackage{color, colortbl}
\usepackage{stfloats}
\usepackage{enumitem}
\usepackage{tabularx}
\usepackage{xstring}
\usepackage{multirow}
\usepackage{xspace}
\usepackage{url}
\usepackage{subcaption}
\usepackage{xcolor}
\usepackage[hang,flushmargin]{footmisc}

\ifcamera \usepackage[accsupp]{axessibility} \fi

% ...

\ifarxiv  \fi

\newcommand{\R}[1]{{%
    \textbf{%
        \ifstrequal{#1}{1}{\textcolor{red}{R#1}}{%
        \ifstrequal{#1}{2}{\textcolor{blue}{R#1}}{%
        \ifstrequal{#1}{3}{\textcolor{magenta}{R#1}}{%
        \ifstrequal{#1}{4}{\textcolor{teal}{R#1}}{%
                           \textcolor{cyan}{R#1}%
        }}}}%
    }%
}}

\usepackage{xr-hyper}

\makeatletter
\newcommand*{\addFileDependency}[1]{
  \typeout{(#1)}
  \@addtofilelist{#1}
  \IfFileExists{#1}{}{\typeout{No file #1.}}
}

\makeatother

\definecolor{cvprblue}{rgb}{0.21,0.49,0.74}
\usepackage[pagebackref,breaklinks,colorlinks,allcolors=cvprblue]{hyperref}
\usepackage[capitalize]{cleveref}
\crefname{section}{Sec.}{Secs.}
\crefname{table}{Table}{Tables}
\crefname{figure}{Fig.}{Figs.}

\ifarxiv \crefname{appendix}{App.}{Apps.}
\else \crefname{appendix}{Suppl.}{Suppls.} \fi

\begin{document}

\title{\paperTitle}
\author{\authorBlock}

\twocolumn[{%
\renewcommand\twocolumn[1][]{#1}%
\maketitle
\begin{center}
    \centering
    \captionsetup{type=figure}
        \includegraphics[width=0.99\linewidth]{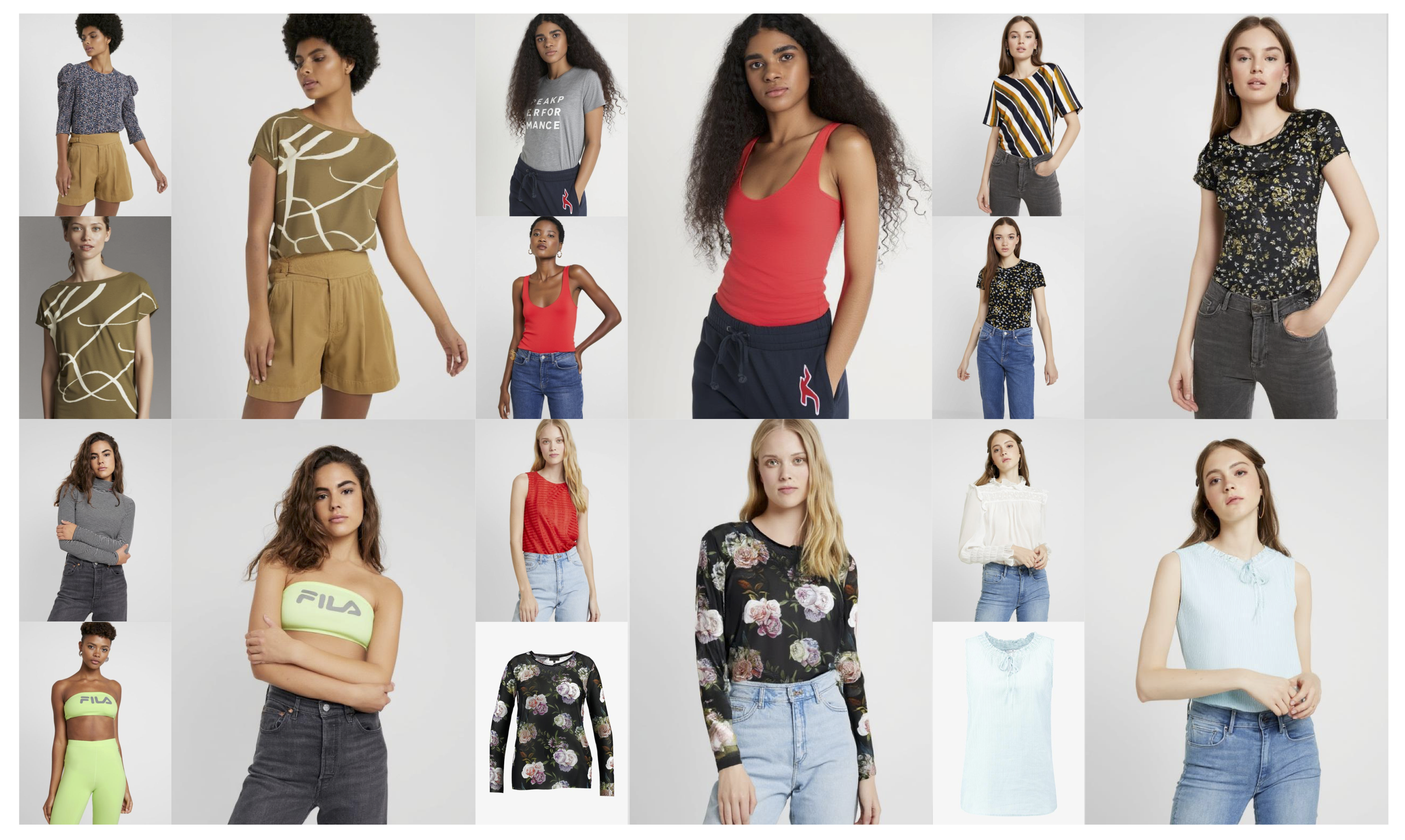}
    \captionof{figure}{Person-to-person and garment-to-person try-on outcomes generated by our MFP-VTON.}
    \label{fig:showresult}
\end{center}%
}]

\begin{abstract}
The garment-to-person virtual try-on (VTON) task, which aims to generate fitting images of a person wearing a reference garment, has made significant strides. However, obtaining a standard garment is often more challenging than using the garment already worn by the person. To improve ease of use, we propose MFP-VTON, a Mask-Free framework for Person-to-Person VTON. Recognizing the scarcity of person-to-person data, we adapt a garment-to-person model and dataset to construct a specialized dataset for this task. Our approach builds upon a pretrained diffusion transformer, leveraging its strong generative capabilities. During mask-free model fine-tuning, we introduce a Focus Attention loss to emphasize the garment of the reference person and the details outside the garment of the target person. Experimental results demonstrate that our model excels in both person-to-person and garment-to-person VTON tasks, generating high-fidelity fitting images.
\end{abstract}

\section{Introduction}
\label{sec:intro}


The initial VTON tasks~\cite{choi2021viton, morelli2022dress, kim2024stableviton, choi2024improving, xu2024ootdiffusion, zhang2024boow} focus on generating realistic fitting images of a person dressed in a reference garment, referred to as garment-to-person in this paper. This task requires the garment to be accurately transferred while ensuring that the person details remains unchanged, except in the area covered by the garment. With the advancement of generation models~\cite{goodfellow2014generative,isola2017image, ho2020denoising, song2020denoising, rombach2022high, esser2024scaling}, this task achieves highly realistic fitting outcomes and find applications in the e-commerce industry~\cite{zalando2024}. 

In the garment-to-person task, the garment is typically a product image, free from obstruction or distortion. However, it is more difficult to obtain compared to the garment already worn by a person. To achieve person-to-person VTON, three main approaches have been proposed. First, some users attempt to segment the garment from the reference image and then use existing state-of-the-art models for try-on. However, this garment misalignment between training and inference often introduces noticeable artifacts. To address this, some researchers~\cite{zeng2020TileGAN, velioglu2024tryoffdiff,xarchakos2024tryoffanyonetiledclothgeneration, tan2024ragdiffusion, shen2024igrimprovingdiffusionmodel} use an auxiliary model to recover the standard garment from the reference person image before applying VTON to generate fitting outcomes, yielding better fidelity than the previous method. However, this approach significantly increases computational load and introduces garment restoration errors. More directly, some researchers~\cite{xie2022pasta,cui2024street} concentrate on developing end-to-end methods for the person-to-person task. However, due to the absence of a suitable dataset, they resort to using unpaired data, which has resulted in limited progress and generated fitting images that still fall short of expectations.

The garment-to-person model has made significant progress in generating high-fidelity fitting images. In this paper, to address the lack of person-to-person datasets, we first use an off-the-shelf garment-to-person model~\cite{choi2024improving} to generate person-to-person paired data. Similar to CatVTON-FLUX~\cite{catvton-flux} (unless otherwise specified, we refer to CatVTON-FLUX simply as CatVTON throughout this paper), we employ Flux-Fill-dev~\cite{flux} as our foundation model. It is built upon diffusion transformers~\cite{peebles2023scalable, esser2024scaling} and introduces advanced inpainting and outpainting capabilities. We find that directly masking the garment region on the target person during generation results in pose inconsistencies between the target person and the final fitting output, as well as the loss of foreground details such as mobile phones held in hands. In the contrary, we directly concatenate the reference garment and the original target person as input condition to preserve as much original information as possible. To avoid compromising the performance of the pretrained model, we also concatenate a full-mask image for inpainting. Additionally, to better guide the model's fine-tuning, we introduce Focus Attention Loss, which helps the attention in transformer prioritize the garment on the reference person and the details outside the garment of the target person.

Our contributions are summarized as follows:
\begin{itemize}
    \item   We propose MFP-VTON, a mask-free VTON model that generates realistic fitting images with the reference garment from product image or worn by another person.
    \item	We prepare a custom dataset for person-to-person VTON, consisting of high-quality person images, and introduce Focus Attention loss to ensure that both reference garment and the details outside the garment of the target person receive full attention.
    \item	We evaluate the superior performance of our proposed method, showcasing its ability to outperform other state-of-the-art approaches in person-to-person task while achieving comparable results in garment-to-person task.
\end{itemize}

\section{Related Work}
\label{sec:related}

\subsection{Virtual Try-On}
Current virtual try-on methods can be categorized into garment-to-person~\cite{choi2021viton, morelli2022dress, kim2024stableviton, choi2024improving, xu2024ootdiffusion, zhang2024boow, catvton-flux} and person-to-person~\cite{xie2022pasta, cui2024street}. The methods corresponding to the former are mostly trained on VTON-HD~\cite{choi2021viton} and DressCode~\cite{morelli2022dress}, which contain high-resolution paired data of standard garments and person images. These methods mask the areas of garment to be generated, indicating where the new garment should be placed, but this leads to the loss of foreground information. In contrast, ~\cite{zhang2024boow} proposes a mask-free method that does not explicitly define the areas to be fitted, helping to preserve the original image details. However, these methods still rely on garments as references. To achieve person-to-person try-on, ~\cite{zeng2020TileGAN, velioglu2024tryoffdiff,xarchakos2024tryoffanyonetiledclothgeneration, tan2024ragdiffusion} suggests using an additional model to restore the standard garment from the person image. Several works~\cite{xie2022pasta,cui2024street} attempt to address person image-based try-on using unpaired datasets due to limitations in the datasets, but the fitting results remain suboptimal. In our work, we use a state-of-the-art VTON model~\cite{choi2024improving} to create a person-to-person dataset and design a mask-free framework based on FLUX-Fill-dev~\cite{flux} capable of handling both garment-to-person and person-to-person tasks.

\section{Method}
\label{sec:method}
\begin{figure*}[ht]
    \centering
    \includegraphics[width=0.99\linewidth]{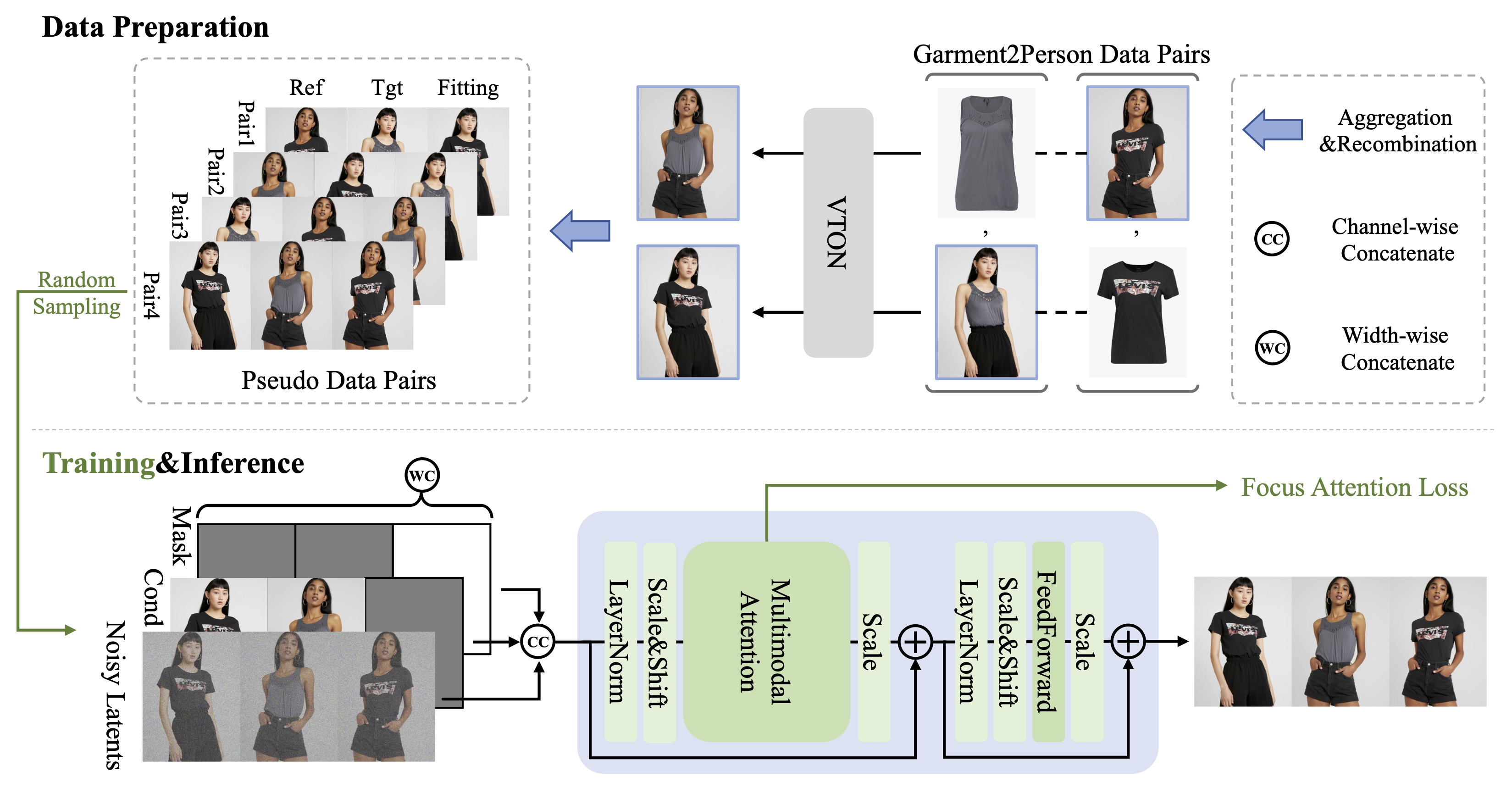}
    \caption{Overview of our proposed method. The upper part illustrates the data preparation process for the person-to-person task. The lower part demonstrates the training and inference pipelines. }
    \label{fig:fig3_model}
\end{figure*}

\subsection{Data preparation}\label{sec:data_preparation}
To build a custom dataset for the person-to-person task, we follow the methodology outlined in~\cite{zhang2024boow}. First, we randomly sample two paired instances from the garment-to-person dataset~\cite{choi2021viton}. Each instance consists of a person image and the corresponding garment, such as $\left(P_{mm},G_{m}\right)$ and $\left(P_{nn},G_{n}\right)$, where $P_{mm}\in\mathbb{R}^{3\times H\times W}$ represents the person $P_m$ wearing the original garment $G_m$. We then use an off-the-shelf VTON model~\cite{choi2024improving} to virtually swap garments between these two persons, generating new data pairs, $\left(P_{mn},G_{n}\right)$ and $\left(P_{nm},G_{m}\right)$, where $P_{mn}$ denotes person $P_m$ wearing the new garment $G_n$. As shown in the top of~\cref{fig:fig3_model}, this process results in four pseudo-data triplets: $\left(P_{nn}, P_{mm}, P_{mn}\right)$, $\left(P_{mm}, P_{nn}, P_{nm}\right)$, $\left(P_{nm}, P_{mn}, P_{mm}\right)$, and $\left(P_{mn}, P_{nm}, P_{nn}\right)$. Each triplet consists of the person wearing the reference garment, the target person to be fitted, and the target person wearing the reference garment, respectively.

\subsection{Mask-Free VTON with Diffusion Transformer}\label{sec:VTON}
We adopt FLUX.1-Fill-dev~\cite{flux} as the foundation model, which has the advanced inpainting capability to fill masked regions in an existing image based on text guidance. The model's input comprises Gaussian noise, a masked condition $cond\in\mathbb{R}^{3\times H\times W}$, and its corresponding mask $M\in\mathbb{R}^{1\times H\times W}$. Here, the condition represents the preserved portion of the image $I\in\mathbb{R}^{3\times H\times W}$ after masking, expressed as $cond=I\cdot\left(1-M\right)$. The output of the model maintains the same dimensions as the condition, with the unmask region remaining unchanged. We draw inspiration from In-Context-LoRA~\cite{huang2024context} by concatenating multiple images directly along the width dimension to form the image condition. To preserve the details of the target person, we avoid masking the garment regions of the target person, in contrast to the approach used in CatVTON-FLUX~\cite{catvton-flux}. For example, using $\left(P_{nn}, P_{mm}, P_{mn}\right)$ as a training example, the reference person $P_{nn}$ and the target person $P_{mm}$ are concatenated directly to create the condition. As shown in the bottom of~\cref{fig:fig3_model}, to align with the foundation model, we append an additional blank region $BLK\in\mathbb{R}^{3\times H\times W}$, which serves as the area for inpainting. Consequently, while the input condition $cond=Concat\left(P_{nn}, P_{mm}, BLK\right)$ is of size $3\times H\times 3W$ and the corresponding mask has the same spatial dimension, the actual inpainting is performed only on the final region of size $3\times H\times W$. This approach eliminates the need for a garment mask on the target person, resulting in a mask-free input condition.

\subsection{Focus Attention Loss}\label{sec:training_loss}
Since the model needs to focus on both the garment areas of the reference image and the regions outside the garment in the target image, guiding it with text alone is challenging. Therefore, we introduce the Focus Attention loss to directly supervise attention. Specifically, when calculating attention in the FLUX-Fill-dev~\cite{flux}, text embeddings $T\in\mathbb{R}^{b\times l_1\times c}$ and image latent $L=Concat\left(G, P, F\right)\in\mathbb{R}^{b\times 3l\times c}$ are concatenated along the spatial dimension, where $G$, $P$ and $F$ represent intermediate features of the reference person, target person and the fitting outcome, respectively. $l$ is the reshaped spatial dimension of $G$, $P$ or $F$ in latent space.
The query and key can be disassembled into $\left[T_x, G_x, P_x,F_x\right]$, where $x$ is in $\{q, k\}$. After computing the attention matrix, we get the components related to the fitting output $F$: $\left[F_qT_k, F_qG_k, F_qP_k, F_qF_k\right]$. Only the two sub-matrices of attention $F_qG_k\in\mathbb{R}^{n\times l}$ and $F_qP_k\in\mathbb{R}^{n\times l}$ are related to the image condition. Therefore, during the fine-tuning of the model, the garment mask $M_r\in\mathbb{R}^{1\times H\times W}$ for the reference image and the garment mask $M_t\in\mathbb{R}^{1\times H\times W}$ for the target image are utilized to compute the loss as follows:
\begin{equation}
\begin{array}{rcl}
\mathcal{L}_{\mathrm{FA}}=\frac{1}{n}\sum_{i=1}^{n}\mathrm{mean}(Attn_{FG\_i}\cdot (1-M_r) \\ +\mathrm{mean}(Attn_{FP\_i}\cdot M_t)), 
\end{array}
\end{equation}
where, attention map $Attn_{FG\_i}\in\mathbb{R}^{1\times l}$ and $Attn_{FP\_i}\in\mathbb{R}^{1\times l}$ represent components of $F_qG_k$ and $F_qP_k$, respectively. Additionally, $M_r$ and $M_t$ are resized and reshaped to match the dimensions of the attention map.

\section{Experiments}
\label{sec:experiment}

\subsection{Experimental Setup}\label{sec:exp_set}
\noindent \textbf{Implementation Details.} 
Our proposed model is fine-tuned on VITON-HD~\cite{choi2021viton}. As with other works~\cite{xu2024ootdiffusion,choi2024improving,velioglu2024tryoffdiff}, we divide it into a training dataset and a testing dataset. Then, we use IDM~\cite{choi2024improving} to prepare the custom datasets for person-to-person task and manually filter out a subset for training. We adopt the FLUX-Fill-dev~\cite{flux} as our foundation model and fine-tuning it on both garment-to-person and person-to-person datasets. In inference stage, the model samples 30 steps to get the final fitting outputs.

\subsection{Qualitative and Quantitative Comparison}\label{sec:exp_comp}
We compare our model with garment-to-person methods OOTD~\cite{xu2024ootdiffusion}, IDM~\cite{choi2024improving}, and CatVTON-FLUX~\cite{catvton-flux}. To adapt these methods for person-to-person tasks, we employ segmentation~\cite{ravi2024sam} and try-off~\cite{velioglu2024tryoffdiff} to extract garment from the reference person. We initially utilize unpaired testing datasets and assess the fidelity of the generated fitting image distributions with three key metrics: FID~\cite{heusel2017gans}, CLIP-FID~\cite{kynkaanniemi2022role} and KID~\cite{binkowski2018demystifying} metrics. In order to more fully evaluate our model, we process the testing dataset using the data preparation method outlined in~\cref{sec:data_preparation} and extract paired datasets such as $\left(P_{mn}, P_{nm}, P_{mm}\right)$ and $\left(P_{nm}, P_{mn}, P_{nn}\right)$. On this dataset, we evaluate the aforementioned metrics and additionally compute SSIM~\cite{wang2004image}, LPIPS~\cite{zhang2018unreasonable} and DISTS~\cite{ding2020image} to evaluate the reconstruction quality between the generated fitting image and corresponding ground truth.

\begin{figure*}[ht]
    \centering
    \includegraphics[width=0.95\linewidth]{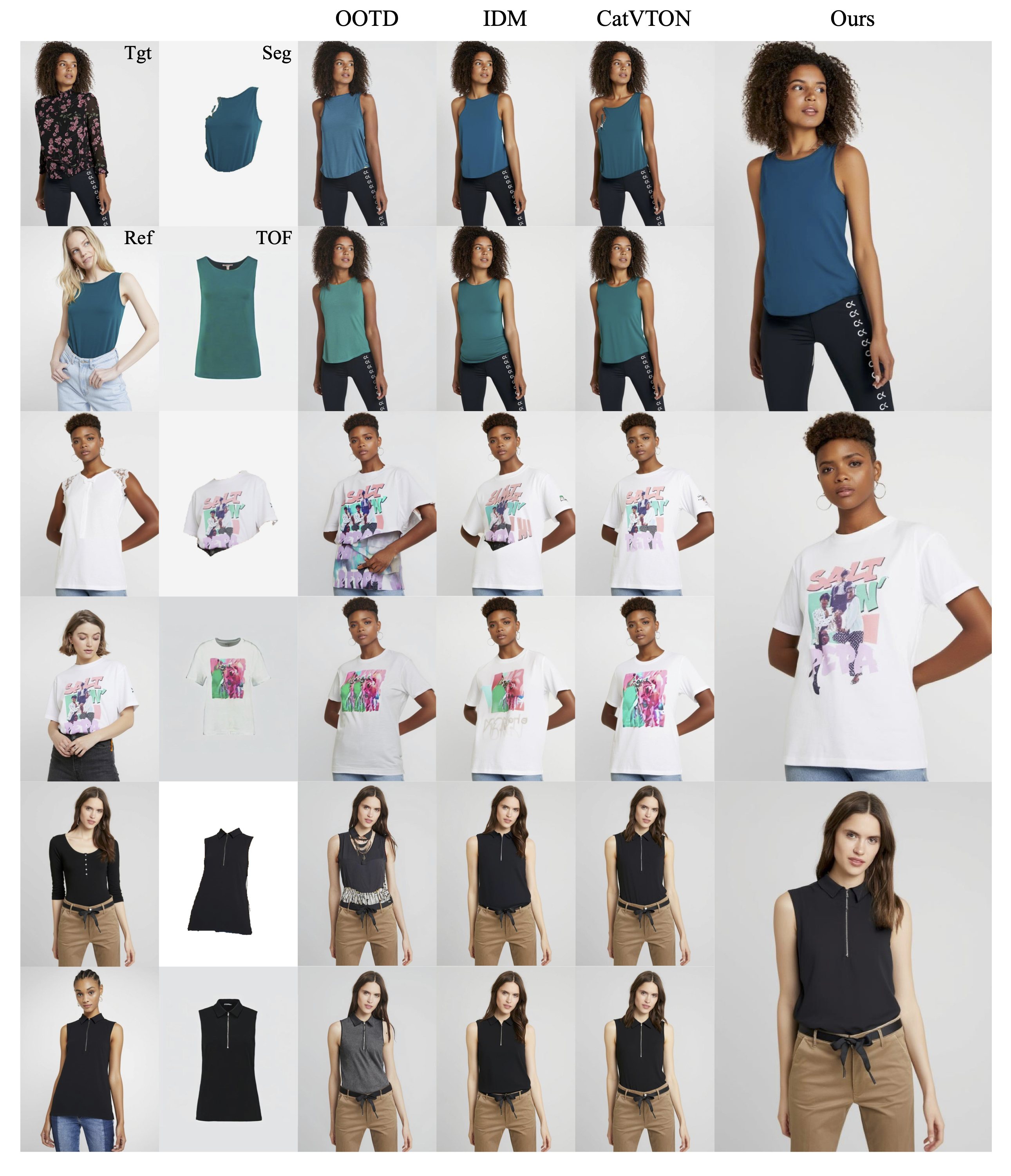}
    \caption{Qualitative comparison. The first two columns show the inputs to different models. In the person-to-person task, the three garment-to-person methods rely on segmentation and try-off techniques to obtain the garment on the reference person. In contrast, our method directly generates the outputs based on the reference person.}
    \label{fig:fig4_method}
\end{figure*}
\noindent \textbf{Qualitative Comparison.}
As illustrated in~\cref{fig:fig4_method}, our method achieves superior fidelity in person-to-person task. While other methods can adapt to person-to-person task using segmentation or try-off techniques, they often introduce significant artifacts. Despite not requiring a separate input of the person pose, our method effectively preserves the original pose with high accuracy.

\begin{table*}[htbp]
\centering
\begin{tabular}{l|cccccc|ccc}
\toprule
\multirow{2}{*}{Model} & \multicolumn{6}{c|}{Paired Person2Person}            & \multicolumn{3}{c}{Unpaired Person2Person} \\ \cmidrule(){2-10} 
                       & SSIM$\uparrow$    & LPIPS$\downarrow$  & DISTS$\downarrow$  & FID$\downarrow$     & CLIP-FID$\downarrow$ & KID*$\downarrow$    & FID$\downarrow$             & CLIP-FID$\downarrow$       & KID*$\downarrow$          \\ \midrule
Seg+OOTD             & 0.8404 & 0.1445 & 0.1081 & 12.4351  & 3.3757 & 3.5754      & 13.3704   & 3.9595        & 4.3530       \\
Seg+IDM              & \underline{0.8727} & 0.1170 & 0.0957 & 11.0887  & 2.6419 & 3.6665      & 10.8623  & \underline{2.6477}       & 3.0886       \\
Seg+CatVTON     & 0.8715 & 0.1150 & \underline{0.0897} & \underline{9.7622}  & 2.9928 & 2.5167      & 10.6096  & 3.0508       & 2.8575       \\ \midrule
TROF+OOTD            & 0.8409 & 0.1368 & 0.1047 & 11.1590  & 3.0541 & \underline{2.1543}     & 11.7932   & 3.5123       & 2.5396       \\
TROF+IDM             & \textbf{0.8761} & \underline{0.1139} & 0.0950 & 10.5302  & \underline{2.5589} & 2.3982      & 11.2508  & 2.7594       & 2.5920      \\
TROF+CatVTON    & 0.8723 & 0.1158 & 0.0923 & 9.8190  & 2.6181 & \textbf{1.9341}       & \underline{10.5839}  & 2.7688        & \textbf{2.3509}       \\  \midrule
Ours                 & 0.8688 & \textbf{0.1122} & \textbf{0.0870} & \textbf{9.3223} & \textbf{2.1333}   & 2.1581  & \textbf{10.3465}         & \textbf{2.2885}        & \underline{2.4658}      \\ \bottomrule
\end{tabular}
\caption{Quantitative comparison with other methods on person-to-person task. The KID metric is multiplied by the factor 1e3 to ensure a similar order of magnitude to the other metrics.}
\label{tab:quantitative_person}
\end{table*}

\begin{table*}[htbp]
\centering
\begin{tabular}{l|cccccc|ccc}
\toprule
\multirow{2}{*}{Model} & \multicolumn{6}{c|}{Paired Garment2Person}            & \multicolumn{3}{c}{Unpaired Garment2Person} \\ \cmidrule(){2-10} 
                       & SSIM$\uparrow$    & LPIPS$\downarrow$  & DISTS$\downarrow$  & FID$\downarrow$     & CLIP-FID$\downarrow$ & KID*$\downarrow$    & FID$\downarrow$             & CLIP-FID$\downarrow$       & KID*$\downarrow$          \\ \midrule
OOTD             & 0.8556 & 0.1118 & 0.0849 & 6.8680  & 2.2030 & \textbf{1.4632}       & 9.8221 & 2.8306 & \textbf{1.6700}      \\
IDM              & \textbf{0.8789} & \textbf{0.0940} & 0.0806 & 6.6752  & 2.1008 & 1.7398   & \underline{9.6548} & \underline{2.4607} & 1.8081       \\
CatVTON     & \underline{0.8774} & \underline{0.0975} & \textbf{0.0776} & \textbf{6.3788}  & 2.2642 & 1.6641      & 9.7696 & 2.7375 & 2.0727      \\ 
Ours                 & 0.8761 & 0.0986 & \underline{0.0790} & \underline{6.4206} & \textbf{1.8431}   & \underline{1.5260}  & \textbf{9.5728} & \textbf{2.2566} & \underline{1.7624}      \\ \bottomrule
\end{tabular}
\caption{Quantitative comparison with other methods on person-to-person task. The KID metric is multiplied by the factor 1e3 to ensure a similar order of magnitude to the other metrics.}
\label{tab:quantitative_garment}
\end{table*}
\noindent \textbf{Quantitative Comparison.}
Quantitative results demonstrate that our method excels in both person-to-person task, as evidenced in~\cref{tab:quantitative_person}, and garment-to-person task, as shown in~\cref{tab:quantitative_garment}, outperforming existing methods across multiple metrics. Additionally, quantitative results indicate that the try-off method is more effective than the segmentation method in facilitating the realization of person-to-person tasks.
\section{Conclusion}
\label{sec:conclusion}
In this paper, we introduce a mask-free framework for person-to-person Virtual Try-On (VTON) based on FLUX-Fill-dev. Given the absence of a dedicated dataset for person-to-person tasks, we construct a custom dataset by leveraging existing garment-to-person model and dataset. Additionally, we propose a Focus Attention loss to enhance attention-level learning, enabling the model to concentrate more effectively on the garment area of the reference person and the outside regions of the garment on target person. Experimental results demonstrate that our method generates high-fidelity fitting outcomes for both garment-to-person and person-to-person tasks.

{\small
\bibliographystyle{ieeenat_fullname}
\bibliography{11_references}
}


\end{document}